# Getting from Generative AI to Trustworthy AI:
# What LLMs might learn from Cyc

Doug Lenat
doug@cyc.com

Gary Marcus
gary.marcus@nyu.edu

July 31, 2023

# Abstract

Generative AI, the most popular current approach to AI, consists of large language models (LLMs) that are trained to produce outputs that are *plausible,* but not necessarily *correct.* Although their abilities are often uncanny, they are lacking in aspects of reasoning, leading LLMs to be less than completely trustworthy. Furthermore, their results tend to be both unpredictable and uninterpretable.

We lay out 16 desiderata for future AI, and discuss an alternative approach to AI which could theoretically address many of the limitations associated with current approaches: AI educated with curated pieces of explicit knowledge and rules of thumb, enabling an inference engine to automatically deduce the logical entailments of all that knowledge. Even long arguments produced this way can be both trustworthy and interpretable, since the full step-by-step line of reasoning is always available, and for each step the provenance of the knowledge used can be documented and audited. There is however a catch: if the logical language is expressive enough to fully represent the meaning of anything we can say in English, then the inference engine runs much too slowly. That's why symbolic AI systems typically settle for some fast but much less expressive logic, such as knowledge graphs. We describe how one AI system, Cyc, has developed ways to overcome that tradeoff and is able to reason in higher order logic in real time.

We suggest that any trustworthy general AI will need to hybridize the approaches, the LLM approach and more formal approach, and lay out a path to realizing that dream.

# 1. Introduction

For all the progress in artificial intelligence in the last decade, trustworthy artificial intelligence remains elusive. Trained statistically with an astronomical number of parameters on large quantities of texts and images, today's AI's (ChatGPT, Bard, etc.) are certainly impressive, but they have been trained to be *plausible*, but not necessarily *correct*. As a result, they are untrustworthy, unstable and brittle. Some examples of what we mean by these terms:

- *Untrustworthy.* After one physician received an unexpected answer from ChatGPT, he asked for the source, for a reference. ChatGPT responded by citing a very convincing-sounding journal article -- but, it later turned out to be a *completely nonexistent* journal article: "*It took a real journal, the European Journal of Internal Medicine. It took the last names and first names… of authors who have published in said journal. And it confabulated out of thin air* [the title of] *a study that would apparently support this viewpoint*". [Faust 2023]  It's fortunate that he not only asked for a citation but also then tried to find and read the cited article. As another example, lawyers were recently sanctioned for citing six nonexistent cases "found" by ChatGPT, and multiple people have been accused of crimes they did not commit, such as a sexual harassment, based on confabulated evidence.
- *Unstable*. As one recent study (Chen et al, 2023) showed, large language models can be quite unstable in behaviors (e.g., determining whether a given integer is prime or composite) from one month to the next.
- *Brittle.* LLM's also sometimes make mistakes that no person would make. E.g., after ChatGPT told us that Romeo commits suicide at the end of *Romeo and Juliet*, we asked whether Romeo dies during the play, and it said there was no way to know! It also answered incorrectly when we asked whether Vladimir Putin believes that cats can breathe, and when we asked whether non-round skateboard wheels would work as well as round ones. Another recent study showed that systems could undermined by adversarial attacks that no human would fall to prey to (Zou, et al, 2023).

In our view, the underlying problem is that LLMs understand too little about the nearly limitless richness of how the world works, about everyday life, human interaction, different cultures, etc. It's notoriously difficult to define "understanding" concisely -- we discuss over a dozen components of "understanding" in the next section. But it mostly comes down to *knowledge, reasoning,* and *world models* (Marcus, 2020), none of which is well handled within Large Language Models.

- ❖ Knowledge: People know many individual *facts*, but equally importantly we have (i) a large, broad, stable base of common sense (e.g., "*you can't be in two places at once*"); and (ii) a large set of qualitative models of how the world works (e.g., "*if it rains, uncovered outdoor items will get wet*").
- ❖ Reasoning: We routinely combine pieces of knowledge and perform multi-step reasoning. If we hear that the President is flying into town tomorrow afternoon, we might adjust our schedule or our planned routes accordingly. If we see on the evening news that the President is making an unplanned trip elsewhere instead, we re-adjust to that. If the US should finally elect our first female president, we would effortlessly generalize that knowledge, regardless of prior history.

In a nutshell, humans possess knowledge and reasoning capabilities, which resemble Kahneman's System 2 (which the second authors calls "deliberative reasoning"), but today's generative AI – more like Kahneman's fast and automatic "System 1" do not. As a result, much of what is obvious to people remains unreliable within the large language model approach.

The next section teases apart "knowledge" and "reasoning", breaking them down into 16 key elements. Section 3 then discusses the progress of accomplishing each of them in a particular "System 2" AI today, Cyc, which is very different from an LLM. Finally, Section 4 considers the ways in which these two types of AI's might work together to produce a trustworthy general AI.

## 2. Sixteen Desiderata for a *Trustworthy* General AI

A general AI which is trustworthy needn't think exactly the same way humans do, but it ought to at least possess the following 16 capabilities:

1. *Explanation.* A trustworthy AI should be able to recount its line of reasoning behind any answer it gives. Asking a series of repeated *Why is that?* follow-up questions should elicit increasingly fundamental knowledge, ultimately bottoming out in first principles and "given" ground truths. Each piece of evidence, knowledge, rule of thumb, etc. invoked in that reasoning chain should also have its source or provenance known. This is a higher standard than people hold each other to, most of the time, but is expected in science and whenever there is a very important decision such as one involving family healthcare, finance, and so on. The explanation should be as concise as appropriate, prioritizing and filtering details based on context and prior and tacit knowledge the user has (or is inferred to have), and resource constraints the user is under (or is inferred to be under).
2. *Deduction*: A trustworthy AI should be able to perform the same types of deductions as people do, as deeply as people generally reason. If you know that countries have borders, and Andorra is a country, then you can infer that Andorra has borders. That use of *modus ponens* is one type of deduction. Another type is arithmetic: if someone enters a room that had four people, it now has five. Exhaustive search is another type of deduction: A chess player soon to be checkmated considers the tree of all moves and counter-moves at that point, and tips over their king. Understanding connectives like *and, or, not* is important, including various "flavors" of negation (e.g., not being able to conclude P is different from being able to conclude P is false.) Deduction also includes recognizing when one statement blatantly contradicts another, and when one statement is obviously redundant with another.

3. *Induction.* Often thought of as a complement to deduction, when certain conclusions cannot be logically deduced. A typical example: An animal's species generally determines the major features of its anatomy. So if you hear about a new type of invertebrate that has just been discovered - let's call it a dwim --- and hear or see that it has eight legs and two wings, you induce that most dwims will have eight legs and two wings.This kind of reasoning sometimes leads to errors but it helps us cope with the rich, complicated world that we live in. A nearly-ubiquitous form of inductive reasoning is *temporal projection:* If you believe or know that X is true at time $t_1$, then you infer how likely it is to be true at time $t_2$. E.g., I learn you own a house, from which I can infer how likely it was you owned it 2 years ago or 3 years from now. Most such projections follow one type of probability decay curve (linear, normal, Gaussian, etc.) for each direction, with the corresponding parameters. Similar projections apply across location, security, and dozens of other dimensions. Things change at boundaries (e.g., state lines) and interrupting events (e.g., getting divorced and selling your house, or less dramatically the ringing of a phone).
4. *Analogy.* Much human reasoning involves analogizing to far-flung and (superficially) unrelated things. The ability to do that by its very nature requires knowing about that vast, broad, panoply of things (objects, actions, properties, etc.) to which one might be analogizing.
5. *Abductive Reasoning,* sometimes known as inference to the best explanation. If a janitor sees a set of chairs in a room that looks like the set of chairs the janitor observed the night before, the presumption, possibly incorrect, but best explanation, other things being equal, is that it is the same set of chairs. This kind of reasoning can lead to errors but, like induction and analogy, it is so useful that we do it all the time.
6. *Theory of Mind:* When we talk with another person, we usually have (or quickly build up) a good model of what they know, are capable of, care about, and so on. We then use that model to guide our interactions: to be more terse with a colleague, to be less terse with a stranger, to use simpler concepts and vocabulary with a young child, etc. Similar presumptions about prior and tacit shared knowledge occur when interacting with your neighbor, with someone who is about your age, with someone who is much older/younger, with someone attending or participating in the same event, etc. An overly loquacious AI could appear condescending, patronizing, or pedantic; one that's too terse could appear cryptic, uncooperative, and, most seriously, frequently be misunderstood. If conversing with a person who is ambiguous or vague, the AI should be able to infer whether, in each instance, it's better (given the conversation "goal") to adopt and reflect that level of vagueness or to ask some clarifying questions or to avoid both of those paths and instead just temporize (delay), e.g., by changing the subject. The AI should revise its model of other agents (and indeed of the world as a whole) over time -- ultimately over the entire lifetime of each person it interacts with -- adding new temporally tagged revisions rather than overwriting and only keeping the latest model around. One channel of information informing its model of person/group/idea X is *in*direct: what others have said about X, taking *their* models into account of course. One other aspect of Theory of Mind worth mentioning is a model of *self:* understanding what it, the AI, is, what it is doing at the moment and why, and -- very importantly -- having a good model of what it does and doesn't know, and a good model of what it is and isn't capable of and what its "contract" with this user currently is (see item 13, below).

7. *Quantifier-fluency:* Consider "*Every Swede has a king*" versus *"Every Swede has a mother"* -- each person in Sweden of course has the same king but not every Swede shares the same mother! In logic such ambiguities are naturally avoided by using *variables* which are *quantified.* The first sentence would be written "*There exists* a king *x* such that *for each* Swede *y*, *x is y's* king"[1] and in the second would be written "*For each* Swede *y*, *there exists* a mother *x* such that *x* is *y's* mother." Of course non-logicians still understand what is meant by each of the syntactically similar English sentences because of their common sense, the mental models they already have about families, motherhood, monarchies, etc.

8. *Modal-fluency.* Besides those two quantifiers, we often qualify statements with phrases like "I *hope that…*". "He *is afraid that…*", "Jane *believes that…*", "Iran *plans to…*", "...so it is *possible* that…", "...so it *must be the case* that…", etc. These pervade human speech and writing, and people are quite good at correctly and almost effortlessly reasoning with such so-called modal operators, including quite deep nestings of such, e.g., "Ukraine hopes that the U.S. believes that Putin plans to…"

9. *Defeasibility.* Much of what one hears, reads, says, believes, and reasons with is only *true by default.*[2]. New information arrives all the time, and many conclusions that were reached would have turned out differently if that new information had been known at the time. To be trustworthy, an AI needs to be able to assimilate new information and revise its earlier beliefs and earlier answers. For some critical applications, it might need to actively inform others that it's retracting and revising some response it gave them in the past.[3]

10. *Pro and Con Arguments.* Many complicated real-world questions don't have a clear objective answer (e.g., Which college should I go to? What car should I buy?) Instead, there are generally a set of pro- and con- arguments for each possible answer. In some cases, they can be weighted and scored. (Often, humans rely on a set of heuristics for *preferring* one argument over another, such as: prefer recent ones to stale ones, short ones to long ones, expert ones to novice ones, constructive ones to nonconstructive[4] ones, etc). Because almost everything we, or an AI, knows is just default-true, even seemingly straightforward questions might have multiple incommensurably-good answers, and each *answer* could have its own set of pro- and con- arguments.

---

[1] More precisely, there exists *exactly one such* king x. Instead of packing that into this axiom (and having to repeat that in stating other axioms), it's cost-effective to factor it out as a separate axiom, a separate rule of thumb: One generally does not have two or more different kings at the same time.

[2] For example, you know the "rule" that each person has a biological mother who is or was a different, and older, person. But even that rule must have some exception(s) or else there would already have been an infinite number of people born! There are a few sorts of things which are absolutely true, with no exceptions, and always will be, such as how to spell the English word "misspell", what the sum of two integers must be, and the rules of chess… well, the rules *today* (the 75-move rule was added in 2014).

[3] People are often hit-or-miss doing this sort of accommodation, and as we age we inevitably retain more and more "stale" conclusions; AIs can do better than people at this, and might usefully serve as a sort of mental prosthesis or amplifier to help us avoid such unwanted remnants.

[4] If there are 367 people in a building, a *nonconstructive* argument can be made that at least two have the same birthday based on the number of days in a year. This doesn't tell us *who* those individuals are.

11. *Contexts.* Some pieces of advice apply at football games but not classroom lectures (e.g., stand up and cheer to signify approval). Some statements are true in someone's (or some group's) belief system, but not in others'. Some, such as who the king of Sweden is, change over time. Having knowledge contextualized, and the ability to reason within and across contexts, can be vital. It is also essential to be able to reason *about* contexts, such as when trying to decide in which context(s) an entailment should or should not be inferred to hold true. Most human communication leaves some elements of the context *implicit,* which can lead to, e.g., conflation when training an LLM. When performing a task (e.g., interacting with a person), the *use context* is important: inferring why they are being asked this, what resource constraints they may be under, what context is the user in, what use will be made of their response, and so on. *Culture* is one important type of context, and making this explicit should reduce miscommunications and facilitate cross-cultural interactions.

12. *Meta-knowledge and meta-reasoning.* A trustworthy reasoner -- be it human or AI -- needs to be able to access and reason about its own knowledge, ideally including the history and provenance[5] of each fact or rule of thumb, and should have an accurate and realistic model of what it does/doesn't know, and how good/bad it is at various tasks. Wild guesses should not be advanced as blithely as those supported by strong arguments. The reasoner may sometimes benefit by pausing, in the midst of working on some problem, to *reflect:* introspect about what tactic it has been trying, how well that seems to be working out, how much longer it's going to require, reason about whether it might be better to change tactics or strategies, and so on. After the fact it might be important to analyze what it did, so as to improve its tactics and strategies for the future. The AI should be able to introspect and explain why it changed its mind about something from yesterday, and hypothesize plausible scenarios which would cause it to change its mind about something -- then cache those, and be alert for signs that they may be occurring. The AI should be able to understand, and come up with, jokes, which usually involve meta-reasoning, common sense, and a theory of mind. Another important type of meta-reasoning is critical thinking about whether and when some particular source can be trusted. Theory of mind, contexts, pro/con argumentation (above) can also all be considered types of meta-knowledge and meta-reasoning.

13. *Explicitly ethical.* A trustworthy AI should follow a core of guiding principles which appear inviolate, such as not lying or causing emotional or physical harm. As everyone knows, however, these are often shaded and complicated and conflicting (e.g., lying to someone so as not to unnecessarily hurt them; a doctor snapping a dislocated shoulder back into place) and ever-changing, requiring meta-reasoning to resolve. There will never be universal agreement on what this core "constitution" should be, so one important aspect is that there will be a large space of multiple *contexts* each inheriting from more general ones and imposing their own revisions. Interactions with the AI, and tasks performed by it, would be situated in such contexts.  One important aspect is that the AI's then-current corpus of ethics be explicitly known and inspectable by everyone it is affecting, and this tenet is one of the very few which should never change -- part of the small, immutable, kernel "contract" between an AI and its users.  Successfully performing certain tasks involving interaction with people may require  the AI to be empathetic (or at least sympathetic and apologetic if it is barred from empathizing by that "contract").  Another aspect of this is that the AI will need to make -- and keep -- promises to each person and group of people it interacts with, subject to their "contracts" -- a common example of this would be not betraying confidences.  As always there would be exceptions, such as if a life were in danger, a subpoena, etc [Taylor Olsen 2023].

---

[5] Since LLMs don't explain their reasoning down to the provenance of the elements used, it's unclear how to tell if one LLM infringed on another. This is relevant, now, given how frighteningly easy it is to fast-follow even the best LLMs [Blain 2023]

14. *Sufficient speed.*  Just like a human being working on a task, an AI needs to be responsive *enough* given the type of problem it is working on.  Some applications require microsecond response times (and therefore cannot be performed by people), some require real-time human dialogue response times (on the order of ¼ second), and it's fine for some other applications to run at slower speeds (e.g., writing a complete 200-page NIH grant proposal).  Human "hardware" is relatively uniform, but of course an AI's speed will drastically depend on the computing hardware running it.

15. *Sufficiently Lingual and Embodied.*  Some applications would be almost impossibly difficult without the performer -- a human or an AI -- being able to converse in natural language, or hear and speak (understanding and generating appropriate prosody), or visually parse scenes and recognize objects, move around, manipulate physical objects, use instruments and devices, sense texture, pressure, temperature, odors, etc.  The modifier "*Sufficiently*" is important, since many applications will require little if any such embodiment, motor or perception capabilities, and little if any natural language dialogue capabilities.  Natural language understanding alone involves many AI-complete elements, such as correct disambiguation, understanding of idioms, metaphor, sarcasm, foreshadowing, irony, subtext, and so on.  Almost all the knowledge and reasoning of the AI is language-independent (e.g., the fact that writing pens are smaller than penitentiaries has nothing to do with the fact that the English string "pen" could denote either one).  But there is still of course the matter of knowing and mastering the lexicon, grammar, idioms, and so on for different natural languages, and mapping the language-specific terms and phrases to terms and expressions in the AI's representation of knowledge.

16. *Broadly and Deeply Knowledgeable.*  We take for granted that anyone we speak or write to has a vast shared foundation of fundamental knowledge about the world, from common sense to models of traffic, weather, crime, etc.  Knowing a vast plethora of *facts* is less important today than it used to be, thanks to the omnipresent internet, and Google in particular, but an effective person or AI needs to be able to access (and understand) facts they need as they need them; humans rely heavily on web searching, AI's may be a little less facile at that but better at accessing databases, web services, and structured websites.  Per the preceding 15 elements, a trustworthy AI should leverage the meaning of each piece of knowledge it has or acquires[6]: be able to explain it, reason about its provenance and trustworthiness, deduce things that logically follow, induce and abduce what a reasonable person would, analogize to (and from) it at least as well as most people do, etc., and do all that as quickly as necessary.

There are some other capabilities which are effectively combinations of the above 16.
- One important one is *Planning.* Planning can involve combining all the various types of reasoning above (deductive and inductive, analogy, meta-reasoning, weighing pro and con arguments, etc.)  The same goes for *Choosing,* as in selecting an item to buy, and in prioritizing tasks.
- Another example is *Learning.*  That also can and should draw on all the above types of reasoning capabilities.  E.g., a typical 2023 robot might require an enormous number of spills and accidents to learn how to clean a hotel room or drive a car, but a human only requires only a modest amount of experience to become proficient, because of all the background knowledge and reasoning skills we have.

Any general artificial intelligence should have these 16 capabilities[7] if it is to be trusted where the cost of error is high.  LLMs today struggle with most of these 16 categories of learning.  In the next section, we discuss how Cyc, perhaps the extreme realization of System 2 (deliberative reasoning) existing on Earth today, approaches these.

# 3. How Cyc handles some of these 16 elements

Large Language Models such as OpenAI's ChatGPT and Google's BARD and Microsoft's Bing/Sydney represent one pole in potential architectural space, in which essentially neither knowledge nor reasoning is explicit.  Cycorp's CYC represents the opposite pole: a four-decade-long 50-person project to explicitly articulate the tens of millions of pieces of common sense and general models of the world that people have, represent those in a form that computers can reason over mechanically, and develop reasoning algorithms which, working together, are able to do that reasoning sufficiently quickly.

Whereas LLMs are trained automatically, statistically, quickly, from large text corpora, usually using self-supervised learning, Cyc has been built by painstakingly identifying each useful axiom individually and then writing it by hand in a logical formalism, and then entering that into the growing Cyc knowledge base.   This process has been accelerated by gamification, NLU, etc., but each axiom is hand-checked for default correctness, generality, and best placement into the microtheories (contexts) it applies to, before entering it into the Cyc knowledge base.

---

[6] We can't count on being as lucky as Chance the gardener in [Kosinski 1971].
[7] There are many additional criteria one could add to the above list (e.g., functioning in a way that faithfully mirrors human cognition including error rates and delay times; or having a sense of humor) but these 16 are the most frequently important ones.  Conversely, there are of course many quite useful applications such as calculator and calendar apps, that lack almost all of those 16 capabilities; we understand when and how and what to trust them with, not unlike the way we treat physical tools.

Cyc began with a frame-and-slots representation akin to today's Knowledge Graphs [Lenat, Prakash, and Shepherd 1986], and an inference engine that ran expert systems style if-then rules, also known as situation-action rules.  Gradually, over its first several years, the indispensability of having an expressive representation language -- one as expressive as English, Arabic, or Portuguese -- became clear.  Namely, a trustworthy general AI needs to be able to represent more or less anything that people say and write to each other -- e.g., "Putin wants the U.S. to be worried that shortly after any U.S. tanks actually arrive in Ukraine, China will blockade Taiwan".

A natural language like English is of course sufficiently expressive to express enormous nuance (though perhaps not, e.g, to express knowledge of what a cat looks or how to use a can opener).  But the AI also needs to be able to algorithmically infer the things that people would, from such statements.  That's one of the main reasons we communicate with other people: not just to have our utterances *remembered*, but also have the listener/reader *reason* with them and, if/when/as appropriate, *act* on them.

Some of that expected reasoning happens right away, inferences we expect our audience to immediately conclude; and some will occur in the future, when what we just said to them might have some impact on their future reasoning and decision-making.   There are various methods that computer science, linguistics, and AI have developed to reason from sentences represented in natural language, such as latent semantic indexing, but those are very incomplete.

A strong alternative to natural languages exists, namely representing each utterance in a formal logical language.  Algorithms can then mechanically operate on sets of such statements, and produce all their entailments automatically, one by one.  There are many such logics to choose from, but only higher order logic can represent the same breadth of thought as a natural language.   By the late 1980's, Cyc assertions and rules were expressed in such a language, CycL.  It includes full first order logic (with variables, nested quantifiers, predicates, functions, etc.), and allows statements *about* other statements, statements about functions, statements about what the inference engine is trying to do at any moment and why, the earlier sentence about Putin and Ukraine, and so on.[Lenat and Guha 1990]

The main reasoning mechanism in Cyc is akin to mechanical "theorem proving" on sentences expressed in predicate calculus -- formal logic.  An example of this is sketched in Figure 1.  The *reasoning* is 100% logically sound, but premises like "people love their children" are only true the vast majority of the time, not in every case, so at the end it is merely extremely likely that the person watching their daughter take her first step is smiling.  That's why we put "theorem proving" in quotes: what gets inferred is not a guaranteed-to-be-true theorem, it's just a strong argument.

> Let's walk through a simple example of first-order machine deduction. Suppose we have a situation like: "A person sees their daughter take her first step." An AGI should be able to answer a question like "Is that person smiling, in that situation? (And, if so, why?)"
>
> The first step in applying machine deduction is to express both the situation and the question in logic. The 3 variables p, d, e, represent the person watching, their daughter, and the walking event. "∧" is the symbol for AND. The negated clause says: there is no *prior* event, f, in which the person d was walking.
>
> **The situation**:
> A0.  (∃p) (∃d) (∃e) is-a(p, Person) ∧ daughter(p, d) ∧ is-a(e, Event) ∧ sees(p, e)
>     ∧ action(e, Walking) ∧ performer(e, d)
>     ∧ (∄ f) (is-a(f, Event) ∧ action(f, Walking) ∧ performer(f, d) ∧ startsBefore(f, e))
>
> **The question**: expressionDuring(p, e, Smiling)     ← True or False: p is smiling during that event
>
> Assume that there is also a set of "common sense" axioms which are available to use in bridging between the situation and the question. In English, six of these would say:
> A1. People love their children.
> A2. If you find out that someone you love has accomplished something significant, it makes you happy.
> A3. When something makes you happy, you smile.
> A4. Taking one's first step is a significant accomplishment for people.
> A5. If you see some event happening, you know the performer and the action
> A6. A person's daughter is one of their children.
> In logic, A1 for example would be: (∀x)(∀y) ((is-a(x,Person) ∧ parent(x,y)) ⇒ loves(x,y)). More literally, this would be read as "For any x and y, if x is a person and the parent of y, then it follows that x loves y."
>
> As mentioned above, these rules of thumb are all just *true by default*. So they can be used to deduce an *argument* for the person p smiling, not a *proof* that that must be the case. To produce that argument, the first step is to *negate* the question:
> NQ:  ¬expressionDuring(p, e, Smiling).     ← "¬" is the symbol for NOT.
> Then, step by step, two of the available axioms get *unified* to produce a new conclusion, a lemma. The available axioms, at any step in this process, include axioms A1-A6, plus the situation A0, plus the negated question NQ, plus all the lemmas derived so far. For example, unifying A6 and A1 produces the conclusion that people love their daughters. Unifying *that* with A0 produces the conclusion: loves(p, d). After about half a dozen more such reasoning steps, a contradiction is derived -- i.e., something and its negation, which can't both be true. But A0, A1,...A6 are given as True, so it must be NQ that's false. Which means Q must be true. So we now have an *argument* that person p was smiling. The entire step-by-step deduction chain is the answer to *Why?* Namely: your daughters are also your children, you love your children so you love your daughters, etc. etc.

**Figure 1.  Using first-order logic to answer a question.** *Notice that this is a different source of power than the statistically-driven operations over large corpora in Large Language models. Still, even logic can lead to an error, especially if the reasoning chain is very long and each step is just <u>usually</u> true. As discussed later in the essay, in principle both Logic and LLMs might serve as a sanity-check on each other: one might be suspicious if an LLM predicts something which has no logical argument supporting it; and, conversely, be suspicious if a logic-based AI deduces something which seems to have few if any instantiations occurring in any human texts ever written.*

Searching for an *algorithm* that could mechanically grind out all the logical entailments of a set of statements has a long and noble history dating back to Aristotle's syllogisms: all men are mortal, Socrates is a man, hence Socrates is mortal. Following Frege, Whitehead, Russell, and other late 19[th] and early 20[th] century philosophers, enormous progress has been made over the last century in implementing logical reasoning engines. However, achieving adequate speed for highly expressive languages remains an unsolved problem.

To run even tolerably fast, most symbolic AI systems today restrict the logical language in which formal statements are expressed, e.g. to knowledge graph or propositional logic which does not allow quantifiers, variables, modals, etc., or to constrained subsets of first order logic (e.g., description logic), which reduces the computational demands at the cost of expressiveness.   Given the arduous nature of the reasoning required (see Figure 1) and the need for tens of millions of general rules of thumb, not just a handful (A1-A6) of them, it is understandable almost all AI researchers and developers have gone in the opposite direction, abandoning or trivializing symbolic representation and reasoning, and instead seeking one or another sort of "free lunch" in the form of perceptrons, multi-layer neural networks and, most recently, LLMs.

But, per our 16 desiderata, numbers 7 and 8 especially, limiting an AI to such a narrow "baby-talk" language would be a huge barrier to it ever becoming a trustworthy general AI.  Even full first-order logic is much less than what people routinely use in writing and talking to each other[8], and falls far short of what a trustworthy AI must be able to handle correctly and expeditiously in almost every real-world application.

For that reason, Cycorp has persevered, unwilling to sacrifice the expressiveness of the logic involved, and its Cyc AI is the culmination of that effort.  Over the past four decades it has developed *engineering solutions* to manage each of the 16 elements described in Section 2.  Some are elegant; others simply required a lot of elbow grease -- e.g., for item 16, Cyc's knowledge base (KB) comprises tens of millions of hand-authored assertions, almost all of which are general "rule of thumb" axioms (most of the "facts" Cyc knows are ones that it can just look up on the internet much as a person would, or access in databases where the schema of the database has been aligned to Cyc's ontology.)

**Let's turn to items on the list of 16, now**.  The final one-- being Knowledgeable -- sounds like a no-brainer, but what knowledge is and isn't cost-effective to hand-axiomatize?  How did Cycorp decide what to enter into the Cyc KB?  Cycorp had its ontologists examine random pieces of text, identifying places where the writer correctly assumed that the reader would disambiguate some polysemous word, some prepositional phrase attachment, multiple pronoun referents, etc. That in turn then gets articulated as a piece of common sense knowledge.  E.g., consider the sentence "*The horse was led into the barn while its head was still wet*". The thing being wet is the horse, not the barn, not the weather, etc.  Change one word -- "*The horse was led into the barn while its **roof** was still wet*" and now the thing being wet is clearly the barn. That leads to the nuggets "horses have heads", "barns don't have heads", "barns have roofs", etc. The ontologist next formalizes this using the language of predicate calculus and the vocabulary of the (growing as needed) Cyc ontology.  But before and after that formalizing, the ontologist will try to generalize the assertion to the point where it still remains default-true.  In this case, a moderately good generalization might be "animals have heads".  The ontologist would also ferret out, and axiomatize that a horse doesn't have two or more heads (that generalizes to *everything* by default, i.e., it is default-true that anything that has a head only has one head), and that two different horses don't share the same head, a horse has the same head for its entire lifetime, the head connects to the body at the top of the neck, the head faces forward, the head is about a tenth the size of the body, etc.  Of course there are exceptions to all those axioms, such as hydras and Cerberus and rare two-headed live births, but each generalization, each rule of thumb, holds true *by default*.  There can be systematic exceptions, and individual exceptions, and exceptions to the exceptions, etc.

---

[8] Here, e.g., is the first sentence in the first international story on CNN.com as the first author typed these words on March 22, 2023:  "*Florida Gov. Ron DeSantis is making a significant shift in tone toward the war in Ukraine, calling Russian President Vladimir Putin a "war criminal" who should be held accountable, in another portion of a Piers Morgan interview teased in the New York Post.*"

Tens of millions of assertions and rules were written and entered into Cyc's KB by hand, but it is important to realize that even just performing *one step* of reasoning, Cyc could generate tens of billions of new conclusions that follow from what it already knows.  In just a few more reasoning steps, Cyc could conclude trillions of trillions of new, default-true statements.  It generally doesn't just metaphorically sit back and ponder things, though, it is asked questions; and, running on a typical laptop today, Cyc can nearly instantaneously answer any of those trillions of trillions of commonsense inferences.   E.g., how many thumbs did Lincoln's maternal grandmother's mother have?   Consider if we ask Cyc why it thinks that Bullwinkle the Moose and Mickey Mouse are not the same individual.  E.g., it might have looked that fact up in some compendium of facts about pairs of individuals (but there would have been more than $10^{20}$ such facts!)  but a better approach would be if it had applied a more general rule like "Mooses and Mice are disjoint".  But even then, Cyc would need to know about $(10^4)^2$ -- on the order of a hundred million -- similar rules to cover even just the 10,000 most common types of living things.  Instead, decades ago the Cyc ontologists pointed Cyc to the Linnaean taxonomy system and added just one single rule to the Cyc KB of the form: For any 2 taxons, if one is not a specialization of the other (through a series of sub-taxon links), assume they are disjoint.  This type of generalization was critical to have the KB-building enterprise take only (!) a few million person-hours of effort rather than a trillion.

To speed up the educating process, the Cyc team developed tools that made use of the existing Cyc KB (and reasoners) to help the ontologists who were introspecting to unearth and formalize nuggets of common sense.  For example, it was important that they *generalize* each nugget before entering into Cyc's knowledge base.  Suppose the original axiom they jot down is "different horses don't share a leg"; a good default-true generalization of that might be "different physical objects don't share physical parts".  Further generalization is questionable -- e.g., many objects *do* share a cost, country of origin, owner, time of creation, etc.  A software tool helps the ontologist semi-automatically walk up the hierarchy of types from "horse" to "physical object", and from "leg" to "physical part".

Another useful Cyc-powered tool calls the ontologist's attention to any existing knowledge Cyc has that appears to contradict this new assertion.  That's usually a good thing to happen, not a bad one: it points the ontologist to tease apart the *contexts* in which each axiom applies -- e.g., the years in which each held true -- and have them each asserted only in the appropriate context (and its specializations).  Even with those Cyc-powered KB-building tools, it has taken a coherent team of logicians and programmers four decades, 2000 person-years, to produce the current Cyc KB.  Cycorp's experiments with larger-sized teams generally showed a net *decrease* in total productivity, due to lack of coherence, deeper reporting chains, and so on.

The Cyc reasoner produces a complete, auditable, step-by-step trace of its chain of reasoning behind each pro- and con- argument it makes, including the full provenance of every fact and rule which was in any way used in each argument.

While natural language *understanding* is an AI-hard problem, natural language *generation* is more straightforward, and Cyc has templates that enable it to produce a passable English translation of anything which is expressed in its formal CycL representation language.  Additional rules help it stitch the series of steps in an argument into a somewhat stilted but readable English paragraph.  As we discuss in the next section, this might be another opportunity for synergy between Cyc and LLMs.

In describing how Cyc has tackled the 16 desiderata, a crucial question is #14: **how is Cyc able to operate sufficiently quickly**, often producing hundred-step-long arguments in seconds across such a huge KB expressed in higher order logic?

As we have already remarked, symbolic AI systems *other than Cyc often* approach speed very differently.  Many limit their KB (which is what led to stove-piped Expert Systems), or they limit the expressiveness of their representation of knowledge, or they limit the types of operations that can be performed on those (i.e., they adopt a more limited, but faster, logic.)  E.g., they choose knowledge graphs or propositional logic which does not allow quantifiers, variables, modals, and so on.

Cyc addresses this by separating the *epistemological* problem -- what does the system know? -- from the *heuristic* problem -- how can it reason efficiently?   Every Cyc assertion is expressed in a nice, clean, expressive, higher order logic language -- the Epistemological Level (EL) language, CycL -- on which, in principle, a general theorem prover could operate.  Slowly.  Very very slowly.  But Cyc also allows multiple redundant representations for each assertion, and in practice it uses multiple redundant, specialized reasoners -- Heuristic Level (HL) modules -- each of which is much faster than general theorem-proving when it applies.

By 1989, Cyc had 20 such high-level reasoners [Lenat&Guha 1990]; today it has over 1,100.
- For example, one fairly general high-level reasoner is able to quickly handle transitive relations, such as "*Is Austin physically located in the Milky Way galaxy?*"  Often, a particular sub-problem will require chasing through dozens of physicallyLocatedIn links if the theorem prover had to operate on those assertions expressed in higher order logic in the EL, in CycL.  But the transitive-reasoning Heuristic-Level module redundantly stores the full closure of each transitive relation, ahead of time.  When Cyc wants the answer to any such question, that reasoner can just look up the answer in one step rather than having the theorem prover search for a long chain (or the absence of such).

- That reasoner was extremely general; a more specific one handles the case where a problem can be represented as $n$ linear equations in $n$ unknowns.
- A fairly narrow Heuristic-Level module recognizes quadratic equations and applies the quadratic formula.
- Another relatively narrow Heuristic-Level module recognizes a chemical equation that needs balancing and calls on a domain-specific algorithm to do that.

When confronted with a problem, all 1,100 reasoners are effectively brought to bear, and the most efficient one which can make progress on it does so, and the process repeats, over and over again, the "conversation" among the 1,100 Heuristic-Level modules continuing until the problem has been solved, or resource bounds have been exceeded (and work suspends on it). In principle (but see footnote) there is always the general resolution theorem prover with its hand raised in the back of the room, so to speak: it *always* thinks it could apply, but it is the last resort to be called on because it always takes so long to return an answer.[9]

When Cyc is applied to a new practical application, it is sometimes the case that even when it gets the right answers, its current battery of reasoners turns out to be unacceptably slow. In that case, the Cyc team shows to the human experts (who are able to perform the task quickly) Cyc's step by step reasoning chain and asks them to introspect and explain to us how they are able to avoid such cumbersome reasoning. The result is often a new special-purpose Heuristic-Level reasoner, possibly with its own new, redundant representation which enables it to run so quickly. This is what happened, e.g., for a chemical reaction application, where a special notation for chemical equations enabled a special-purpose algorithm to balance them quickly.

The trap the Cyc team fell into was assuming that there would be just one representation for knowledge, in which case it would have to be $n^{th}$-order predicate calculus (HOL) with modals, because it is the only one expressive enough for all AGI reasoning purposes. Committing to that meant vainly searching for some fast general-purpose reasoning algorithm over HOL, which probably doesn't exist. To escape from the trap the Cyc team built up a huge arsenal of redundant representations and redundant reasoners, such that in any given situation one of the efficient reasoners is usually able to operate on one of those representations and make some progress toward a solution. The entire arsenal is then brought to bear again, recursively, until the original problem has been fully dealt with or given up on. That last point raises another aspect of how Cyc reasons quickly: it budgets resources, depending on the application (e.g., acceptable wait times during a conversation with a person), and interrupts reasoners who exceeded their bid on how long they would take, and simply won't bother calling on reasoners who know they will take too long.

---

[9] Note written by the first author: "Something we don't often talk about: We noticed empirically that the general theorem-proving reasoner actually took so long that over a million queries in a row that called on it, as a last resort, just timed out. Going back farther, we saw that that had happened for decades. So, about one decade ago, we quietly turned the general theorem prover off, so it never gets called on! The only impact is that Cyc sometimes runs a bit faster, since it no longer has that attractive but useless nuisance available to it."

**One important aspect that permeates Cyc is *context* -- item 11 on our list.** Philosophers [Barnes 1972] and AI researchers [McCarthy and Hayes 1969] have long advocated for introducing an extra argument in every logical assertion, to stand for the context in which that assertion holds true. In this tradition, Cyc makes each such context a first-class term in its language. Thus one can make assertions and rules *about* contexts, such as the CanadianProfessionalHockey context is a specialization of the CanadianSports, ProfessionalSports, Hockey, Post1900, and RealWorld contexts. Just as many Cyc rules pertain to devices, emotions, everyday actions, and so on, some Cyc rules pertain to -- and reason about -- contexts. One of the most important is knowing, if P is true in one context, C1, and P implies Q is true in another context, C2, then in which contexts C3 is it reasonable to expect Q to be true?

Each Cyc context, also called a Microtheory, has a set of domain assumptions which can be thought of as conjuncts for each of the assertions in that context. Because the common assumptions are factored out, the assertions in that context turn out to be much terser, and the reasoners are thereby able to operate very efficiently within the same context. E.g., consider the 2023 context: every assertion and rule doesn't need to start out "In the year 2023,...". This is even more critical since most contexts are *rich objects:* there is an infinite number of things one *could* say about them. Consider a narrow context like an auto accident. There is no end to the things we *could* assert about it: the color of the hair of each driver, the brand of paint that got smudged, etc. etc. etc.

Asserting two statements in the same context means we have factored out all those shared details, some explicitly (e.g., the date and time of the car accident) the vast majority of which are not worth stating (the position of each blade of grass at the scene) and will never be stated. There are contexts for locations, times, cultures, performers, activities, and contexts for what a person or a group of people believe to be true[10]. Cyc can reason within the StarWars context, e.g., and name several Jedi, and not have a contradiction with the same question being asked in the RealWorld context and answering that there are no Jedi. Today there are about 10,000 contexts explicitly given names in Cyc's ontology; that number is kept *down* thanks to Cyc functions which return contexts as their value, such as IntersectContexts, which obviate the need for reifying combinatorially more contexts.

**#9, Defeasible.** Almost all knowledge in Cyc's KB is merely true-by-default. Exceptions can be stated explicitly, for an individual or for an entire *type.* E.g., teachers today are typically sedentary, but gym teachers are an exception, but this particular person is an exception to that exception, but while recuperating… etc. What this means is that Cyc typically can find multiple "proofs" for an answer, and even multiple "disproofs" for the same answer. Those are in quotes because these are really just alternative lines of reasoning, pro- and con- arguments. That means that Cyc reasoning is based around argumentation, not proof, as the next point explains.

---

[10] In general these are *counterfactual* beliefs from the point of view of a broader context, or else we wouldn't have needed to create that more specific belief context.

Page 15                    © 2023 Doug Lenat and Gary Marcus           From Generative AI to Trustworthy AI

***#11, Pro- and Con- Arguments.***  When asked a question, Cyc gathers all the pro- and con- arguments it can find for all the answers that have at least one pro- or con- argument, and then applies meta-level rules to decide which arguments to prefer over which others.  E.g., more specific ones trump more general ones.  If most Spaniards primarily speak Spanish at home, but most residents of Vitoria primarily speak Basque, and Fred lives in Vitoria, then there is an *argument* that Fred primarily speaks Spanish at home (he's a Spaniard) but a *preferred* argument that he primarily speaks Basque.  There is also a weak but very general argument Fred *doesn't* speak Basque at home, namely that comparatively few people speak Basque.  But the argument that he does speak Basque is preferred to that weak argument.

# 4. Synergizing an LLM and CYC

The two types of AI's have different strengths and weaknesses.

- While Cyc's KB is both deep and broad, it is often not deep and broad *enough;*  while Cyc's natural language understanding and generation is good, it is often not good *enough,* certainly not as good as ChatGPT or BARD which is able to chat (for at least a sequence of a few back and forth utterances) about, well, almost anything.  And while Cyc can reason quickly, it is often not fast *enough,* certainly not as fast as those LLM-trained chatbots who are able to always respond at acceptable human conversation speeds.

- The current LLM-based chatbots aren't so much understanding and inferring as remembering and espousing.  They do astoundingly well at some things, but there is room for improvement in most of the 16 capabilities listed in Section 2, other than breadth and speed and what we might call "the wisdom of the crowd" type reasoning.

How could a system like Cyc help ameliorate this?  More symmetrically, how could a knowledge-rich, reasoning-rich symbolic system like Cyc and an LLM work together, so as to be better than either can on its own?  We see several opportunities for such synergy, some very short-term and some longer-term ones:

1. **Symbolic systems such as Cyc as a Source of Trust, to reject false confabulations**

    LLMs present narratives that are so well-spoken they can make compelling statements that are actually false.  Much of their content is true, of course, since the patterns of language reflect the real world.  But it's a cloudy mirror, due to superficial falsehoods like analogies and metaphors, and less excusable ones like misinformation and disinformation.  Cyc and LLMs might synergize by acting as devil's advocates for each

other, e.g. by asking each to explain the negation of what the other says. BARD is already doing a version of this, by offering to call on Google immediately afterwards to help correct some of what it might have just gotten wrong. Bard's continued confabulations are a reminder however, of how nontrivial this kind of integration is; such an integration should be an important focus of research.

2. **Symbolic systems such as Cyc as a Source of Truth, to bias LLMs towards correctness**

   LLMs are trained on (some fraction of) the corpora of ten trillion sentences on the internet. But there is so much tacit common sense about the world that is assumed yet rarely or never explicitly expressed. For example, search for "Do people believe that cats can breathe?" and you'll get tens of millions of hits that answer *related* questions, but not that one, because *of course* people believe that! And everyone knows that everyone knows that. So why bother ever saying/writing it? Cyc is a repository of default-true common sense, including things like that (of course axiomatized much more generally than about cats and breathing!), which are so fundamental that it would be confusing or insulting for a person to explicitly say or write them when communicating with another human being.

   Cyc's inference capabilities can be viewed as a compression algorithm allowing us to implicitly embed exponentially more default-true statements than what is explicitly stated in the KB. Asking it to proactively reason forward and generate *millions or billlions of default-true statements* that could serve as the basis for training future LLMs to be more biased toward common sense and correctness.

   One could also use envision using Cyc, if it were enhanced with more robust natural language understanding, to filter out inputs to an LLM, inputs that Cyc infers to be false, before the LLM is exposed to and trained on those falsehoods. In that way, the falsehoods never even get into the model. Even better, Cyc might then be able to "fix up" the input, e.g., by more properly contextualizing it, into something which *is* true.

3. **LLMs as generators of candidate assertions/rules to add to symbolic systems such as Cyc**

   LLMs can be asked to translate natural language sentences into CycL, and they already are able to do that! Here is what happened when GPT-3 was asked to write a CycL sentence that means

   *"Did you touch a blue object located in the capital of France on September 25th, 2022?"*

   It responded:
   ```
   (thereExists ?object
    (and
     (isa ?object BlueObject)
     (located ?object (CityFn ParisFrance))
     (thereExists ?date
       (and
         (isa ?date Date)
         (dayOfDate  ?date  25)
         (monthOfDate  ?date  9)
         (yearOfDate  ?date  2022)
   ```

<pre><code>                   (touched ?object ?date)))))
</code></pre>

This at first glance amazed the Cyc team -- it looked for a moment like it might be close to correct. But it turns out to have several serious mistakes and garblings, such as the thing that is touching the blue object is the *date;* and several predicate names are wrong, have the wrong argument number and type, and so on; and there is no referent for "you" in the original English sentence. But it might be close *enough,* it might be cost-effective to have Cyc do the translation starting from that garbled-CycL (plus the original English sentence), rather than starting only with the original English sentence. I.e., it might turn out to be much easier to get Cyc to understand English this way rather than trying to get Cyc to transform open-ended English sentences all on its own, in one step.

This leads to a few additional, potentially game-changing ideas:

3.a. The first is that we could ask the LLM to translate each sentence it was trained on, and/or can generate (obviously starting with relatively tiny subsets of such) into CycL, one sentence or paragraph at a time. Then, given the fractured-CycL sentence(s) suggested by the LLM, we use Cyc to (as automatically as possible) turn it into "real" CycL and contextualize it. During the process, we repeatedly have Cyc ask itself: Can you already prove this? Can you already disprove this? Can existing content (in Cyc) be strengthened to accommodate this? And so on.

3.b. The second idea here is based around the fact that Cyc already has a good CycL-to-English translation (NLG) capability. We could ask an LLM to translate into Cyc the generated English for each CycL assertion that Cyc already knows, i.e., each assertion which is already in the Cyc KB and/or which Cyc can infer (obviously starting with relatively tiny subsets of such). Since we already have a good CycL translation of that sentence -- it's what we started with! -- we could then *correct* the LLM by giving it what we know to be a good CycL translation. This should make the process described in the previous paragraph, 3.a., increasingly accurate, increasingly correct.

3.c. These two processes can be interleaved, and a third process would then be to augment Cyc's NLG so that it generates better, more natural-sounding English sentences. This would initially involve human editors who decide what's "better", and Cyc-fluent knowledge engineers who then tweak and add to Cyc's set of NLG templates to get Cyc to produce the better English translations. Eventually, it's possible that the LLM might get sufficiently proficient at translating CycL into English that Cyc's current NLG utility, and this process 3.c, would become unnecessary. The other way this could go would be for the LLM's ability to generate differently-worded English sentences that mean the same thing, to incrementally improve Cyc's NLG capabilities[11], gradually giving it more and more variety and naturalness in the English that Cyc generates.

4. **Symbolic systems such as Cyc as a Source of Inference to dramatically extend LLM coverage**

---

[11] The simplest, but still useful, example of this would be when the LLM generates a sentence using some English word that wasn't even known to Cyc at that time.

LLMs are capable of recapitulating statements or extending patterns and analogies but both of those are impoverished compared to the novel multi-step inferences that humans can produce from a set of "given" statements.

So the idea here is: Given a set of related statements, e.g., the recent N things that were said in this LLM dialogue with this user, have Cyc work to infer (deduce, induce, abduce, analogize, etc.) new consequences, feed those to the LLM, have it (or Cyc) translate those into English.

The Cyc reasoner therefore could act as a potentially exponential amplifier over the latent content already expressible by the LLM. This would naturally follow once the previous two synergy capabilities, (2) and (3), above, are realized.

LLMs usually contain a "feedforward" layer to help them generalize from their input text to a higher level of abstraction. Cyc could use its understanding of the input text to add a *semantic* feedforward layer, thereby extending what the LLM is trained on, and further biasing the LLM toward truth and logical entailment. (WolframAlpha's extension to ChatGPT is somewhat in this spirit.)

5. **Symbolic systems such as Cyc as Source of Explanation to provide audit and provenance support**

    LLMs' inability to soundly explain how they came to their conclusions and what actual trustworthy (or not) knowledge is the basis for those intermediate steps renders them unsuitable for large classes of applications that require soundness, trust, and auditability, such as medical decision-making applications. Cyc can provide exactly this; provenance and explicit justification are the superpower of machine reasoning over non-symbolic LLM representation.

    Another example of this kind of synergy occurred in a project Cycorp and the Cleveland Clinic did for NIH's National Library of Medicine. Databases have been built up of patients' mapped genomes and the disease that brought them into the hospital. Using that, one can statistically learn correlations between point mutations in their DNA and their disease. But such A→Z correlations turned out to be enormously noisy. Enter Cyc, which took each of those *hypotheses* and tried to find a long chain of reasoning that could account for that (e.g., this mutation is next to this gene, which when expressed would be this protein, which in turn could catalyze this reaction, which… ten steps later interfered with bone resorption, which is why the patient developed early-onset osteoporosis.) Such causal pathways generally made predictions along the way (e.g., that the patient would also have slightly elevated bioactive vitamin-D levels) which could then be independently confirmed or disconfirmed in the patient database. Asking Cyc to explain a particular A→Z correlation

was exponentially faster than just asking it to ruminate and propose plausible causal chains in an unguided fashion.

In conclusion, there have been two very different types of AI's being developed for literally generations, and each of them is advanced enough now to be applied -- and each *is* being applied -- on its own; but there are opportunities for the two types to work together, perhaps in conjunction with other advances in probabilistic reasoning and working with incomplete knowledge, moving us one step further toward a general AI which is worthy of our trust.